\documentclass[11pt,letterpaper,logo,dvipsnames,table]{thuair}

\usepackage{hyperref}
\usepackage{url}
\usepackage{graphicx}
\usepackage{xspace}
\usepackage{enumitem} % Add this package to your preamble
\usepackage{lipsum} % For dummy text to show alignment
\usepackage{lipsum} % For dummy text to show alignment
\usepackage{tcolorbox}
\usepackage{natbib}
\usepackage{caption}
\usepackage{subcaption}

\newtcolorbox{AIbox}[1]{colback=gray!10!white,colframe=black!80!white,title=#1}

\usepackage{booktabs}
\usepackage{multirow}
\usepackage{tabularx} % 或者 array, makecell
\usepackage{calc}

\usepackage{array}
\newcolumntype{C}[1]{>{\centering\let\newline\\\arraybackslash\hspace{0pt}}p{#1}}

\hypersetup{
    colorlinks = true,
    citecolor = {blue},
}

\newcommand{\name}{BudgetThinker\xspace} % name of the toolcandidates: BudgetGuide, TokenTracker, ThinkMeter\newcommand{\xx}{\textcolor{red}{??}\xspace}  % placeholder

\newcommand{\ie}{\textit{i}.\textit{e}.~}
\newcommand{\eg}{\textit{e}.\textit{g}.~}
\renewcommand{\today}{Preprint}

\title{\name: Empowering Budget-Aware LLM Reasoning with Control Tokens}

\author{
Hao Wen$^{1,*}$, Xinrui Wu$^{1,*,\ddagger}$, Yi Sun$^{1}$, Feifei Zhang$^{1}$, Liye Chen$^{1,\ddagger}$, Jie Wang$^{1,\ddagger}$, Yunxin Liu$^{1}$, Yunhao Liu$^{2}$, \mbox{Ya-Qin Zhang}$^{1}$, Yuanchun Li$^{1,\dagger}$ \\
$^{1}$Institute for AI Industry Research (AIR), Tsinghua University \\
$^{2}$Global Innovation Exchange \& Department of Automation, Tsinghua University \\~\\
$^*$Equal contribution. \\
$^\dagger$Corresponding author: Yuanchun Li (liyuanchun@air.tsinghua.edu.cn). \\
$^\ddagger$Work done during internships at AIR, Tsinghua University.\\
\textbf{Source Code}: \url{https://github.com/MobileLLM/BudgetThinker} \\
}

\begin{abstract}
Recent advancements in Large Language Models (LLMs) have leveraged increased test-time computation to enhance reasoning capabilities, a strategy that, while effective, incurs significant latency and resource costs, limiting their applicability in real-world time-constrained or cost-sensitive scenarios. This paper introduces \name, a novel framework designed to empower LLMs with budget-aware reasoning, enabling precise control over the length of their thought processes. We propose a methodology that periodically inserts special control tokens during inference to continuously inform the model of its remaining token budget. This approach is coupled with a comprehensive two-stage training pipeline, beginning with Supervised Fine-Tuning (SFT) to familiarize the model with budget constraints, followed by a curriculum-based Reinforcement Learning (RL) phase that utilizes a length-aware reward function to optimize for both accuracy and budget adherence. We demonstrate that \name significantly surpasses strong baselines in maintaining performance across a variety of reasoning budgets on challenging mathematical benchmarks. Our method provides a scalable and effective solution for developing efficient and controllable LLM reasoning, making advanced models more practical for deployment in resource-constrained and real-time environments. 
\end{abstract}

\begin{document}

\maketitle

% \begin{figure*}[htbp]
%   \centering
%   \includegraphics[width=0.9\linewidth]{figs/intro.png}
%   % \vspace{-8mm}
%   \caption{(1) Illustrations of \name. (2) Parallel scaling law for \name based on DeepScaleR-1.5B \citep{deepscaler2025}. ``Token budget'' denotes the total token number for each reasoning path in our method. \wh{Fake data as placeholder, need revise}}
%   \label{fig:main}
% \end{figure*}

\section{Introduction}
% Recent progress in Large Language Models (LLMs) has been fundamentally driven by \textbf{test-time scaling}, a strategy that allocates more computational resources during inference to let models ``think longer'' before producing an answer \citep{deepseekr1, OpenAI2024o1}. This approach encourages LLMs to explore diverse problem-solving strategies and has unlocked significant performance gains across various domains, including advanced math \citep{deepseekr1, gemini25}, coding \citep{hui2024qwen2, yang2025qwen3technicalreport}, agentic tasks \citep{kimiteam2025kimik2openagentic}, and multimodal reasoning \citep{reasonrft, vlmr1}. 

% However, this strategy of ``thinking longer'' comes at a steep price in terms of computational resources and latency \citep{chen2025think23overthinkingo1like}. This high cost severely limits the deployment of these powerful models in real-time applications, such as vehicle control, robotics, and AI agents demand real-time responsiveness. This creates a clear need for a \textbf{controllable Chain-of-Thought (CoT) process}, where a model can complete its reasoning and provide an answer within a computational budget constrained by the application (like response time within 10 Hz for a robot brain) or by user. This can be viewed as another instruction following capability of LLM which we call \textbf{budget adherence}. 

A key recent progress in Large Language Models (LLMs) is test-time scaling, a paradigm where models are prompted to ``think longer'' by generating extended Chain-of-Thought (CoT) reasoning before delivering a final answer \citep{deepseekr1, OpenAI2024o1}. While this approach has unlocked state-of-the-art performance in complex domains like advanced mathematics \citep{deepseekr1, gemini25} and coding \citep{hui2024qwen2, yang2025qwen3technicalreport}, it comes at the significant cost of increased latency and computational overhead. This reliance on increasingly long reasoning makes simple scaling strategies impractical for real-world systems with strict performance constraints, such as vehicle control, robotics, and AI agents.

% This creates a clear need for a \textbf{controllable CoT reasoning}, where a model can complete its reasoning and provide an answer within a computational budget constrained by the application (\eg under 100ms for autonomous driving \citep{ad} or Flying Drones \citep{chen2023typefly}) or by the user tolerance (\eg for long-term LLM agents, calling LLMs will cost too much overhead). This can be viewed as another instruction following capability of LLM which we call \textbf{budget adherence}. 

This trade-off between performance and efficiency highlights a critical need for \textbf{controllable CoT reasoning}, where a model can tailor its reasoning process to conclude within a specified computational budget. Such a budget could be dictated by an application's hard limits (\eg, under 100ms for autonomous driving \citep{ad, chen2025timelynet} or drone navigation \citep{chen2023typefly}) or by user tolerance for cost and delay in long-running agentic systems. We frame this crucial instruction-following capability as \textbf{budget following}.

Existing approaches to control the reasoning budget, however, have proven insufficient. Directly inserting budget constraints into prompts \citep{pu2025thoughtterminator, Sun2025TimesUp} often fails to reliably control output length \citep{tokenaware}. Other methods that toggle between "thinking" and "non-thinking" modes \citep{yang2025qwen3technicalreport, openai2025gpt5} lack the fine-grained control necessary for variable budgets. Even more advanced training-based techniques using Supervised Fine-Tuning (SFT) and Reinforcement Learning (RL) \citep{hou2025thinkprunepruninglongchainofthought, l1, wu2025lapo} struggle to enforce strict adherence to a specified constraint, a limitation our work confirms and aims to address.

\begin{figure*}[htbp]
  \centering
  \includegraphics[width=0.9\linewidth]{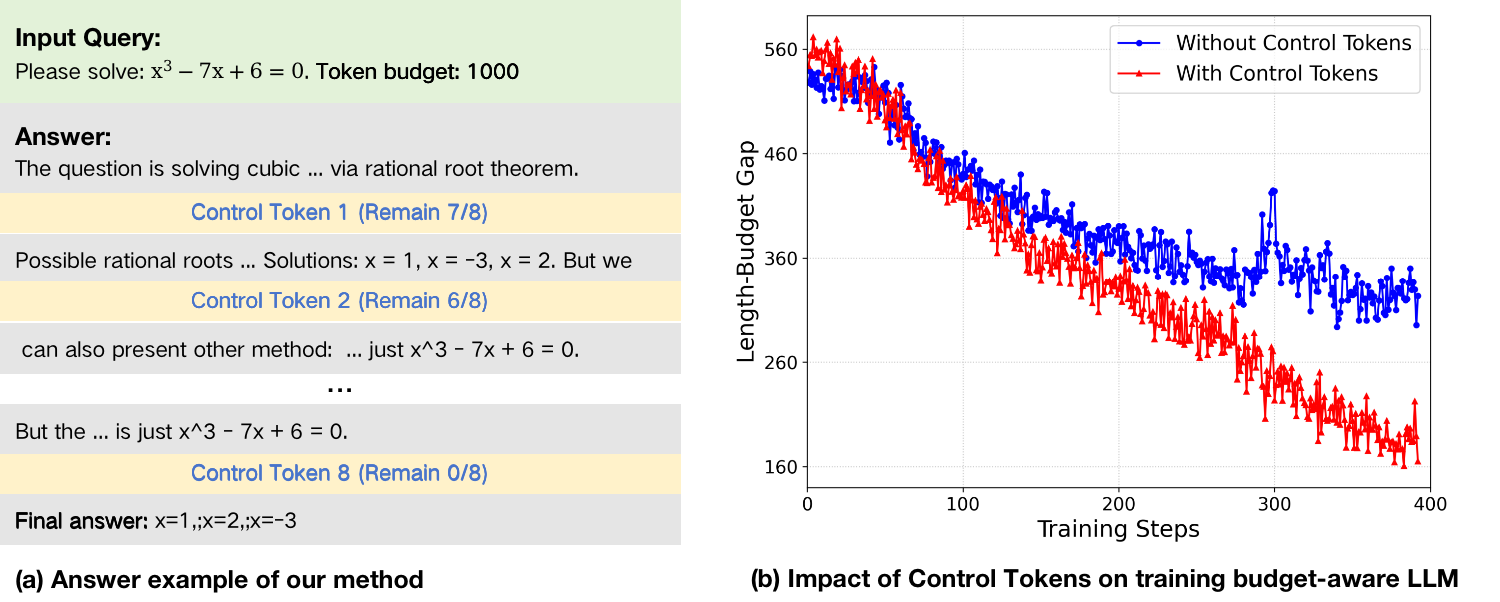}
  % \vspace{-8mm}
  \caption{Illustration of the \name framework and the impact of Control Tokens. (a) Example of budget-aware generation: Shows how BudgetThinker inserts control tokens into its reasoning process to stay aware of the remaining token budget.
  (b) Effect of Control Tokens on training: The gap between the generated answer length and the target budget during reinforcement learning. }
  % \vspace{-8mm}
  \label{fig:intro}
\end{figure*}

In this paper, we address the challenge of precisely controlling the length of a model's thought process. We argue that merely stating a budget constraint in the initial prompt is insufficient, and the model needs to be continuously reminded of its remaining token budget as it generates its response. To achieve this, we develop a novel training and inference framework that enables precise control over the length of the CoT. Our method periodically inserts special \textbf{Control Tokens} that act as explicit signals of the remaining budget, as illustrated in Figure~\ref{fig:intro}a. As shown in Figure~\ref{fig:intro}b, training with these control tokens leads to a faster and more stable reduction in the gap between the generated length and the target budget, demonstrating their effectiveness in teaching budget adherence.

%In summary, our main contributions are:
Our main contributions are:
\begin{itemize}
    \item We introduce a method that inserts special tokens during the inference process to achieve controllable output length. This technique minimizes the number of inserted tokens and their impact on the model's output quality.
    \item We design a comprehensive training process, progressing from SFT to RL, which incorporates a length-aware reward function. This allows the model to adapt to control via special tokens and produce higher-quality, budget-aware outputs. Our method is designed as a plug-and-play module that complements existing test-time scaling training procedures.
    \item We demonstrate that our approach surpasses strong baselines across multiple foundational models and datasets under various reasoning budgets.
\end{itemize}

We evaluate our method based on DeepSeek distilled Qwen 2.5 1.5B and 7B \citep{deepseekr1}, on challenging reasoning benchmarks including MATH-500 \citep{hendrycksmath2021}, AMC 2023, and AIME 2024. Compared to baselines such as the original reasoning models and a state-of-the-art efficient reasoning method \citep{hou2025thinkprunepruninglongchainofthought}, our approach demonstrates a better trade-off between accuracy and token budget. Specifically, our method improves accuracy by an average of 4.9\% across all tested budgets and exhibits more precise adherence to the specified constraints.

% \input{tex/motivation}
% \section{Observations}
% \wh{State the observations that traditional budget-controlling methods couldn't achieve precise control}

\section{Special-Token Conditioned Budget-Aware Reasoning}

We present \name, a budget-aware generation framework that guides LLMs to follow a predefined token budget by conditioning them on explicit control signals. Our method enhances budget adherence by (i) introducing a dynamic control token insertion strategy that continuously reminds the model of the remaining budget and (ii) developing a two-stage training pipeline combining supervised fine-tuning (SFT) and reinforcement learning (RL) to teach the model to follow these control tokens.

\subsection{Special-Token Budget Signaling}

Our method enforces a target output length, or budget, by augmenting the standard autoregressive generation process. The core idea is to \emph{insert a small, fixed set of special control tokens into the sequence at positions corresponding to fractions of the total budget}. These tokens act as explicit signals to the model, indicating how much of its budget has been consumed.
Let $B$ be the target budget and $K$ be a fixed hyperparameter for the number of control signals. We introduce a set of $K$ special control tokens, $\mathcal{C} = \{c_1, c_2, \dots, c_K\}$, where $c_k$ signals that the generation has progressed through the $k$-th fraction of the total budget (\eg, $c_1$ signifies 10\% completion if $K$=10). 
% The length of each interval is then $I = \lfloor B/K \rfloor$. 
% The generation of the token $y_t$ at each timestep $t \in \{0, 1, \dots, B-1\}$ is formally defined by the piecewise function:
During generation, the token $y_t$ at timestep $t$ is determined by the following piecewise function:
$$
y_t =
\begin{cases}
\boldsymbol{c_{k+1}} & \text{if } t = k\cdot \lfloor B/K \rfloor \text{ for an integer } k \in \{0, 1, \dots, K-1\} \\
y'_t \mid (y'_t \sim \pi_{\theta}(\cdot | x, y_{<t})) & \text{otherwise}
\end{cases}
$$
At each timestep $t$ that corresponds to the end of a budget fraction $\lfloor k \cdot B/K \rfloor$, we deterministically insert the corresponding control token $c_k$. For all other timesteps, the next token $y'_t$ is sampled from the LLM's distribution $\pi_{\theta}$, conditioned on the full history of both model-generated tokens and our inserted control tokens. 

For comparison, we define a naive baseline: \textit{Fixed-Interval Signaling}. This simpler strategy inserts control tokens at constant, fixed intervals rather than at budget ratios. Given a fixed interval length $I$, a control token is inserted every $I$ steps (\ie, when $t=k \cdot I$). However, this approach lacks scalability with respect to the budget $B$. For example, if $I=100$, a budget of $B=500$ would require 5 control tokens, while a budget of $B=1000$ would require 10. The model must be trained to recognize a potentially unbounded number of control tokens as the budget grows. In contrast, our ratio-based method always uses the same fixed set of $K$ tokens, making it far more robust to varying budgets.

\subsection{SFT Data Curation}
\label{method:data}

To teach the LLM the semantics of our control tokens, we construct a specialized dataset for Supervised Fine-Tuning (SFT). The curation process involves two key steps: budget assignment and dataset balancing.

\textbf{Budget Assignment. }
First, for each original data sample $(x,y)$, we assign a budget $B$ by rounding the answer's length, $|y|$, up to the nearest multiple of a granularity parameter $T$:

\[
B = T \cdot \left\lceil \tfrac{|y|}{T} \right\rceil,
\]

This calculation method is intentional. By design, the true answer length $|y|$ is almost always less than the allocated budget $|B|$. This gap teaches the model to robustly terminate its reasoning and stop generating tokens once the answer is complete, rather than artificially padding the output to fill the entire budget. 
Next, we transform each sample in the dataset. The input prompt is modified to include the budget constraint, and the target output is reconstructed to include the control tokens:
% \[
% \begin{cases}
% \hat x=\text{Concat}(x, \text{Please answer within $B_i$ tokens)}
% \\
% \hat y=\text{Insert}(y, \{c_j\}_{j=1}^K)
% \end{cases}
% \]
\[
\hat x=\text{Concat}(x, \text{Please answer within $B_i$ tokens)}
\]
\[
\hat y=\text{Insert}(y, \{c_j\}_{j=1}^K)
\]
Here, \emph{Insert} function adds the control token $c_{k+1}$ at each position $t$ when $t=k\cdot I$. 
Then, we get a reconstructed dataset $\{(\hat x, \hat y)\}$, which we use to fine-tune LLM. 

\textbf{Dataset Balancing. }
We balance the dataset to ensure the model's budget following skill is generalizable across different task types and output lengths. We achieve this by creating a mixture of two data categories: 
\textbf{(1) Complex Reasoning Data}: Samples sourced from datasets like s1k \citep{muennighoff2025s1}, LIMO \citep{ye2025limo}, and Bespoke-Stratos-17k \citep{bespoke_stratos}, which typically require long, step-by-step thought processes and generate lengthy responses. 
\textbf{(2) Short CoT Data}: Samples from datasets like NuminaMath \citep{numina_math_datasets} and MATH training set \citep{hendrycksmath2021}, where tasks often have short chain-of-thought extended reasoning (most answers less than 1000 tokens). 
By blending these sources, we ensure our SFT dataset has a well-balanced distribution of target answer lengths. 

% \textbf{Length Rejection Sampling. }

\subsection{Reinforcement Learning}
Following supervised fine-tuning (SFT), we employ Group Relative Policy Optimization (GRPO) algorithm \citep{shao2024deepseekmathpushinglimitsmathematical} to train the model to generate responses that adhere to a specific token budget, $B$. 

\textbf{Reward Design. }
We design a composite reward function that balances factual accuracy with a penalty for deviating from the target length. For a generated response $y$, with ground truth $y_{gold}$ and budget $B$, the reward $R$ is defined as: 

\begin{equation*}
    R(y, y_{gold}, B) = = k_1 \cdot \mathbf{1}\{y = y_{\text{gold}}\} 
    + k_2 \cdot \mathbf{1}\{\text{format}(y)\} 
    + k_3\cdot\underbrace{\max(1 - \gamma\cdot\left(\frac{B - |y|}{B}\right)^2, 0)}_{\text{Length Reward}}
\end{equation*}
% \begin{equation*}
%     R(y, y_{gold}, B) = k_1\cdot r_{answer}
%     + k_2\cdot r_{format}
%     + k_3\cdot\underbrace{\max(1 - \gamma\cdot\left(\frac{B - |y|}{B}\right)^2, 0)}_{\text{Length Reward}}
% \end{equation*}

Indicator functions $\mathbf{1}\{y = y_{\text{gold}}\}$ and $\mathbf{1}\{\text{format}(y)\} $ are the original correctness and format check of GRPO, and $k_1, k_2, k_3$ are coefficients that control the relative importance of each part. 
The \emph{Length Reward} incentivizes the model to use the budget efficiently. It is based on \emph{normalized length deviation}, $\delta = \frac{||y| - B|}{B}$, which measures the fractional difference between the generated length $|y|$ and the target budget $B$. This deviation is modulated by an \emph{asymmetric penalty coefficient}, $\gamma$:

$$
\gamma =
\begin{cases}
1 & \text{if } |y| \le B \\
r & \text{if } |y| > B
\end{cases}
$$

Here, $r > 1$ is a hyperparameter that imposes a substantially larger penalty for exceeding the budget than for falling short of it. This design encourages the model to generate responses that are not only correct but also precisely tailored to the length constraint.

\textbf{Iterative Reinforcement Learning. }
To ensure the model performs robustly across a wide range of budgets, we implement a iterative learning strategy within the RL pipeline. Our approach is based on the hypothesis that generating concise, complete answers under tight budget constraints is more difficult than generating longer, more verbose ones. The iterative, therefore, progresses from easier to harder tasks.
The training is structured in sequential stages. In each stage $k$, the model is trained to generate responses within a specific budget $B_k$. The curriculum proceeds by systematically decreasing the budget across stages, following the sequence $B_1 > B_2 > \dots > B_n$. 

Finally, to prevent catastrophic forgetting and maintain proficiency across all learned budget levels, we conclude with a \textbf{mixed-budget training phase}. In this final stage, a budget $B_k$ is randomly sampled from the set of all previously used budgets $\{B_j\}_{j=1}^n$ for each training batch. This compels the model to retain its ability to generate high-quality responses for any length within the learned spectrum.

\section{Experiments}
\label{sec:exp}
In this section, we present a comprehensive empirical evaluation of \name. We investigate its performance under various computational budgets and analyze its ability to adhere to these budgets (Section~\ref{sec:main}). We then analyze the impact of different control token insertion strategies (Section~\ref{sec:insertion}), iterative training (Section~\ref{sec:exp_iterative}), and reinforcement learning of \name (Section~\ref{sec:exp_rl}).

\subsection{Experiment Setup}

\textbf{Training Details. }
Our experiments use  DeepSeek-R1 \citep{deepseekr1} distilled Qwen-2.5-1.5B and Qwen-2.5-7B \citep{hui2024qwen2} as backbone models. 
For Supervised Fine-Tuning (SFT), we create a dataset of 41k problem-solution pairs as described in Section~\ref{method:data}. 
Initially, we select all 1k reasoning samples from s1k \citep{muennighoff2025s1}, 0.8k reasoning samples from LIMO \citep{ye2025limo}, and 17k reasoning samples from Bespoke-Stratos-17k \citep{bespoke_stratos}, and a subset from NuminaMath\citep{numina_math_datasets}, Math\citep{hendrycksmath2021}. 
Due to computational resource constraints, we remove all question-answer pairs with Chain-of-Thought (CoT) lengths over 10,000 tokens. 
We train all SFT models for 6 epochs with an initial learning rate of 2e-5 and a maximum context length of 12000. 
% For Supervised Fine-Tuning (SFT), we curate a dataset as described in Section~\ref{method:data}. Initially, we select all 1k reasoning samples from s1k \citep{muennighoff2025s1}, 0.8k reasoning samples from LIMO \citep{ye2025limo}, and 17k reasoning samples from Bespoke-Stratos-17k \citep{bespoke_stratos}. Additionally, we include a subset(30k reasoning samples) of the NuminaMath\citep{numina_math_datasets} and Math\citep{hendrycksmath2021} dataset, which consists of problems that appear in both OpenThoughts-114k\citep{openthoughts} and NuminaMath-CoT + MATH datasets. Each problem in this subset contains one short and one long chain of thought from the original and OT datasets, respectively. 
% Due to computational resource constraints, we remove all question-answer pairs with Chain-of-Thought (CoT) lengths over 10,000 tokens. Ultimately, we retain 41,926 question-answer pairs in our dataset. 
% We train all SFT models for 6 epochs with an initial learning rate of 2e-5 and a maximum context length of 12000. 

For Reinforcement Learning (RL), we employ the Group Relative Policy Optimization (GRPO) algorithm \citep{shao2024deepseekmathpushinglimitsmathematical}. The training data is a set of 14k math problems, comprising two parts: one part from the math\_splits category of the prm800k dataset \citep{openai_prm800k}, and the other part from the numina\_amc\_aime labeled data in the PRIME-RL/Eurus-2-RL-Data\citep{prime-rl} dataset. Models are trained for 10 epochs with GRPO hyperparameters $\alpha$ and $\beta$ set to 0.01 and 0.01, respectively. The maximum context length during RL is 10000.

We configure \name by setting the number of control intervals to $K=8$. This corresponds to inserting 8 special tokens during decoding, indicating remaining budget fractions of $7/8, 6/8, \dots, 1/8$. The granularity parameter for generating training budgets is set to $T=50$. The importance hyperparameters are set to $k_1=0.7,k_2=0.15,k_3=0.15$. The length reward penalty coefficient $\gamma$ is set as: 
$$
\gamma =
\begin{cases}
1 & \text{if } |y| \le B \\
16 & \text{if } |y| > B
\end{cases}
$$
Namely, the length reward will decade to 0 when the length of answer $y$ exceeds the budget $B$ by 1/4. 
For the iterative training curriculum, we start from a budget of $6000$, then reducing it to $4000$, $3000$, and finally $2000$. Following this curriculum, we conduct a final mixed-budget training phase by randomly sampling $B \in \{6000, 4000, 3000, 2000\}$.

\textbf{Evaluation Details. }
We evaluate our methods and baselines on three challenging math benchmarks: MATH-500 \citep{hendrycksmath2021}, AIME 2024, and AMC 2023. To assess budget control, we evaluate performance across a range of reasoning budgets, from $B=500$ to $B=10000$ tokens. 
For inference, we utilize a modified version of the vLLM engine \citep{vllm}. Our modified engine enforces the budget by truncating the reasoning process once the length limit is reached and appending the ``\texttt{</think>**Final Answer**}'' tags to signal the model for a final answer. 
We then allocate an additional 50 tokens for the final answer generation before terminating the process entirely. 
For each test question, we generate $N$ candidate solutions using sampling with a temperature of 0 and a top-p value of 1 for MATH-500, and with a temperature of 0.6 and a top-p value of 0.95 for AIME 2024 and AMC 2023. We report the pass@1 accuracy. We set the number of samples $N$ to 1 for MATH-500, and to 64 for AIME 2024 and AMC 2023.

\textbf{Baselines. }
We benchmark \name against two baselines: 
\textbf{(1) ThinkPrune} \citep{hou2025thinkprunepruninglongchainofthought}: A state-of-the-art efficient reasoning framework that also uses GRPO for budget control but does not incorporate explicit control tokens. As the official ThinkPrune repository does not provide a 7B model, we limit our comparison to the 1.5B model scale.
\textbf{(2) Original Models}: The DeepSeek-R1 distilled Qwen-2.5-1.5B and 7B models that serve as the foundation for our fine-tuning.

\subsection{Main Performance}
\label{sec:main}

\begin{figure}[htbp]
    \centering
    \begin{subfigure}{0.439\textwidth}
        \centering
        \includegraphics[width=\linewidth]{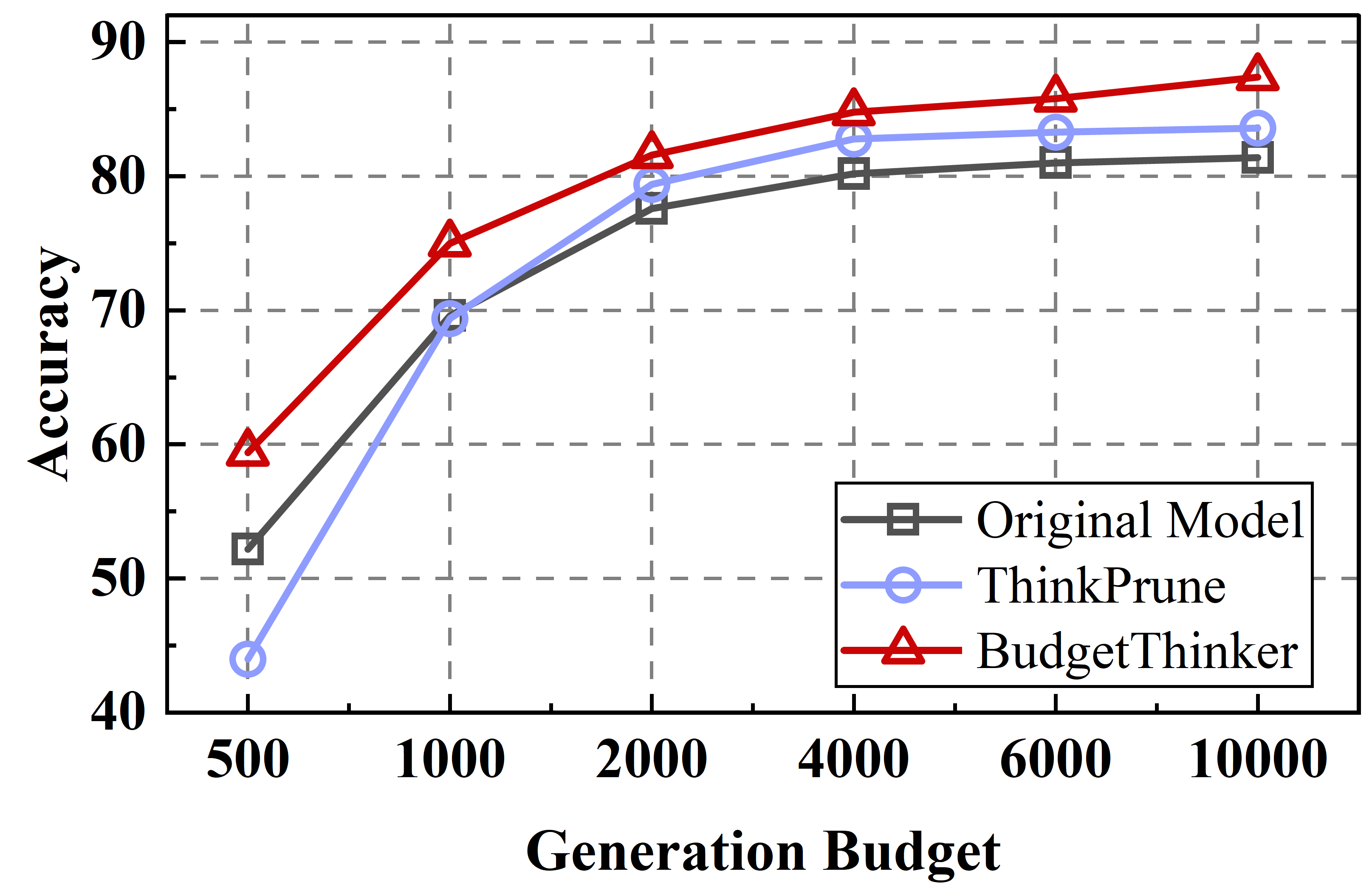}
        \caption{1.5B LLMs on MATH-500}
        \label{fig:scaling_bottleneck}
    \end{subfigure}
    \hspace{0.002\textwidth}
    \begin{subfigure}{0.441\textwidth}
        \centering
        \includegraphics[width=\linewidth]{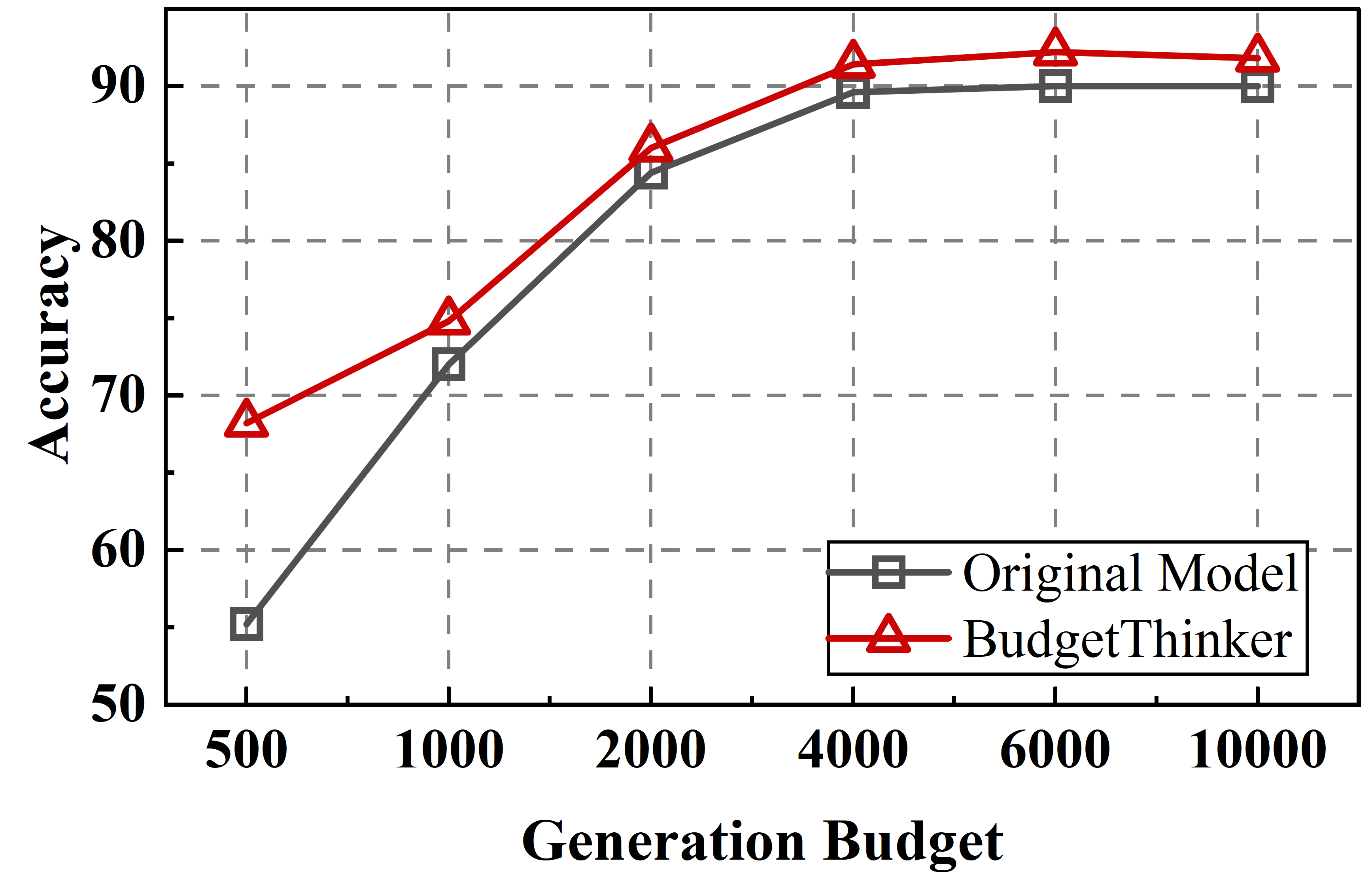}
        \caption{7B LLMs on MATH-500}
        \label{fig:scaling_bottleneck}
    \end{subfigure}
    \begin{subfigure}{0.432\textwidth}
        \centering
        \includegraphics[width=\linewidth]{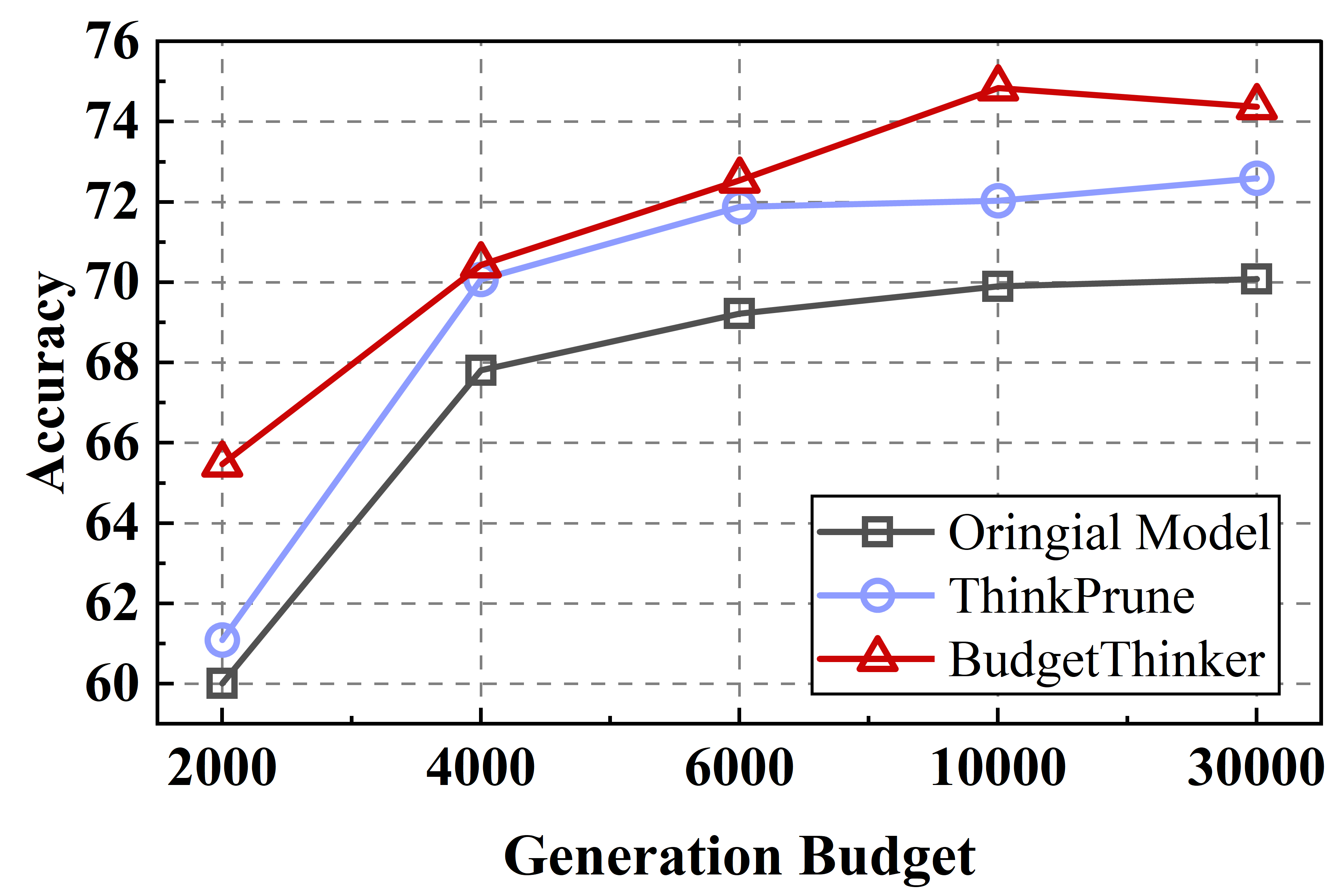}
        \caption{1.5B LLMs on AMC 2023}
        \label{fig:scaling_bottleneck}
    \end{subfigure}
    \hspace{0.0027\textwidth}
    \begin{subfigure}{0.438\textwidth}
        \centering
        \includegraphics[width=\linewidth]{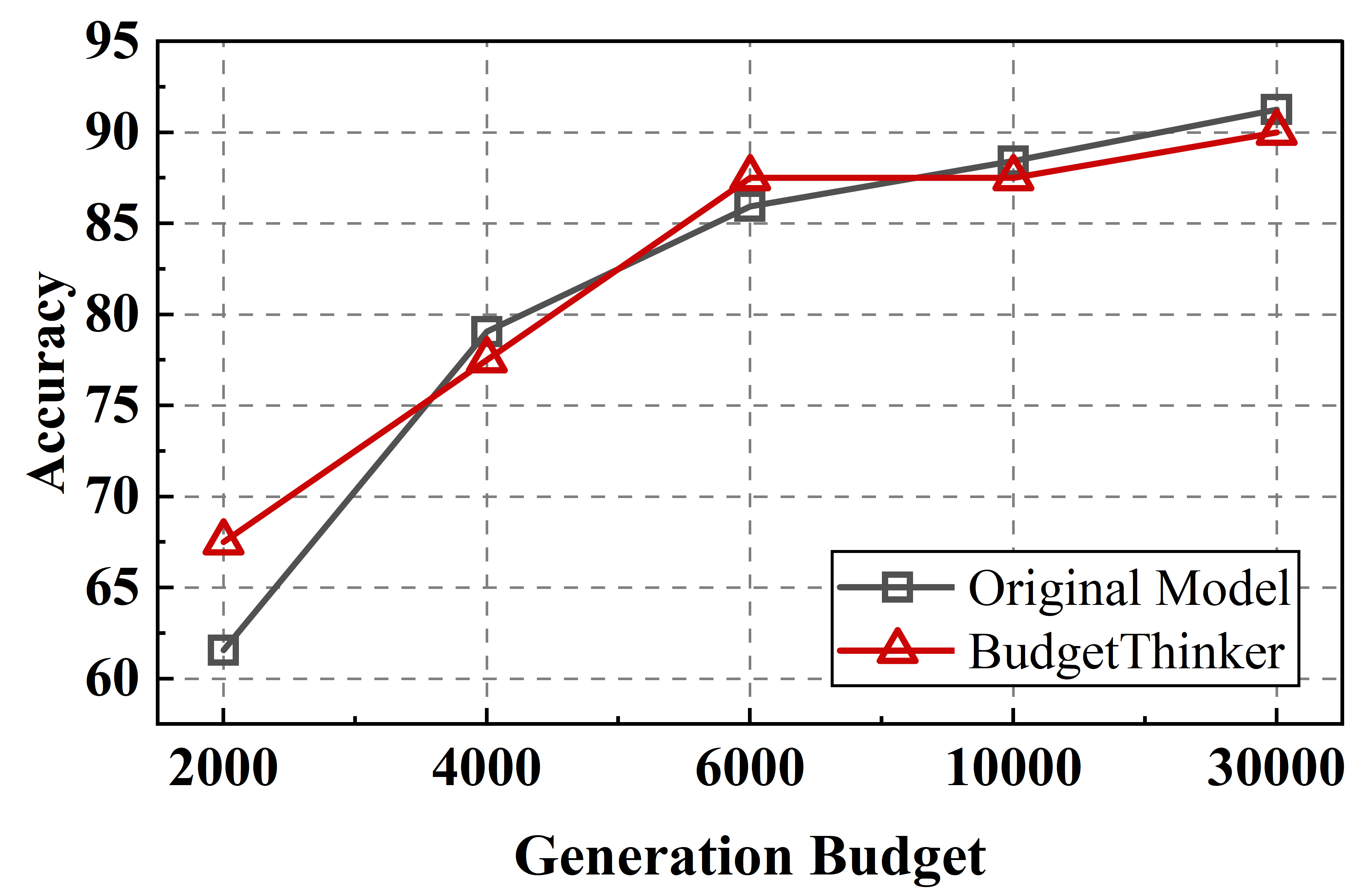}
        \caption{7B LLMs on AMC 2023}
        \label{fig:scaling_bottleneck}
    \end{subfigure}
    \caption{Pass@1 accuracy of BudgetThinker vs. baselines across various generation budgets on MATH-500 and AMC 2023. Accuracy is plotted against the maximum generation budget, a metric more applicable to real-time scenarios than average reasoning length. }
    \label{fig:math_amc}
\end{figure}
\begin{figure}[htbp]
    \centering
    \begin{subfigure}{0.437\textwidth}
        \centering
        \includegraphics[width=\linewidth]{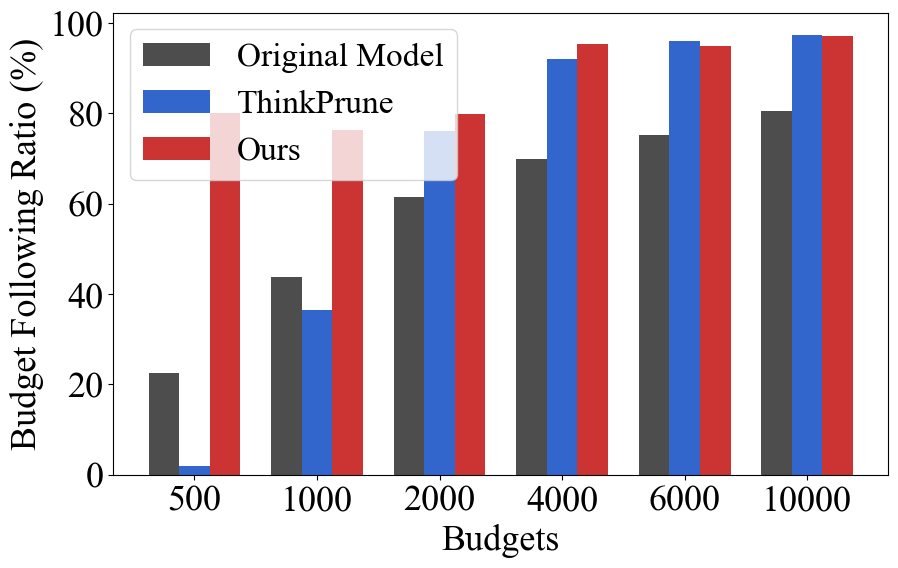}
        \caption{Budget following ratio.  }
        \label{fig:faithful}
    \end{subfigure}
    \hspace{0.002\textwidth}
    \begin{subfigure}{0.437\textwidth}
        \centering
        \includegraphics[width=\linewidth]{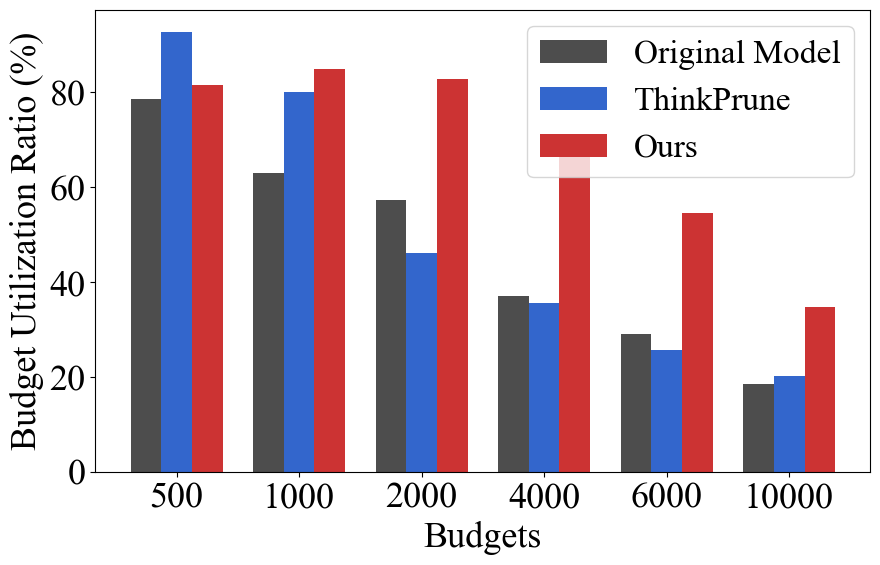}
        \caption{Budget utilization ratio.  }
        \label{fig:utilization}
    \end{subfigure}
    \caption{Budget following and utilization analysis on MATH-500. We compare \name-1.5B, ThinkPrune-1.5B, and the original DeepSeek-Distilled-Qwen-2.5-1.5B on MATH-500. (a): The percentage of generated responses that terminate naturally within the specified budget limit ($B$). (b): For responses that finish within the budget, this shows the average generated length as a percentage of the total allocated budget ($|y|/B$). This shows \name has a superior ability to follow budget constraints while also making more effective use of the allocated tokens compared to baselines.}
    \label{fig:budget_follow}
\end{figure}
We first evaluate the accuracy of \name against the baselines across various token budgets, with results shown in Figure~\ref{fig:math_amc} and Table~\ref{tab:aime_results}. For results of MATH-500 and AMC 2023, \name outperforms the original model and ThinkPrune across most budget allocations. Specifically, \name improves accuracy by 4.2\% over the original model and 5.7\% over ThinkPrune on average. 
However, on the AIME 2024 benchmark, the performance of all three methods was comparable. This is likely because the inherent difficulty of AIME problems demands longer, more complex chains of thought; simply restricting the generation length does not necessarily yield better solutions, as producing a concise yet correct proof requires a higher level of reasoning.
% However, for AIME 2024, the accuracies of all the three methods are similar. This is because the difficulty of these problems requires longer chain of thoughts at the source, and restricting the LLM generation length doesn't produce better solutions, and producing a very concise step requires higher intelligence. 
% This demonstrates that the explicit budget-awareness instilled by our control tokens allows the model to allocate its computational resources more effectively, leading to superior reasoning performance for a given cost.

We also analyze the budget-following capabilities of each model, as illustrated in Figure~\ref{fig:faithful}. 
Besides, in Figure \ref{fig:utilization}, we calculate the average relative length at each budget ($|y|/B$) for every answer that is within budget, which shows the capability for LLM to understand and fully use budgets. 
The original model, lacking any budget-specific training, exhibits poor capability to follow the specified limits. While ThinkPrune \citep{hou2025thinkprunepruninglongchainofthought} shows improved budget awareness, it often underutilizes the allocated budget, prematurely concluding its reasoning process, which can negatively impact performance on complex problems. In contrast, \name demonstrates superior budget adherence. Guided by the remaining budget tokens, it effectively utilizes the entire allocated length without significant overruns. This precise control allows it to achieve a better balance between computational cost and accuracy, leading to its state-of-the-art performance.
\begin{table}[t!]
\centering
\small
\begin{tabular}{l|c|c|c|c}
\toprule
\textbf{} & $B$=2000 & $B$=4000 & $B$=6000 & $B$=10000 \\
\midrule
\textit{Original-1.5B}&       14.80&	23.28&	25.71&	28.88\\
\textit{ThinkPrune-1.5B}& 15.22&	22.78&	25.22&	27.80\\
\textbf{\name-1.5B}&      16.25&	23.80&	25.63&	28.90\\
% \textit{Base-1.5B}& 28.5 & 40.0 & 49.6 & 50.4 \\
% \textit{ThinkPrune-1.5B}& 35.5 & 45.0 & 54.2 & 55.8 \\
% \textit{\name-1.5B}& 24.4 & 35.8 & 42.7 & 45.2 \\
% \textit{Base-7B}& - & - & - & - \\
% \textit{\name-7B}& - & - & - & - \\
\bottomrule
\end{tabular}
\caption{Performance of \name and baselines on AIME 2024 with different budgets ($B$). }
\label{tab:aime_results}
\end{table}

\subsection{Analysis of Control Token Insertion Strategies}
\label{sec:insertion}

\begin{figure}[htbp]
    \centering
    \begin{subfigure}{0.435\textwidth}
        \centering
        \includegraphics[width=\linewidth]{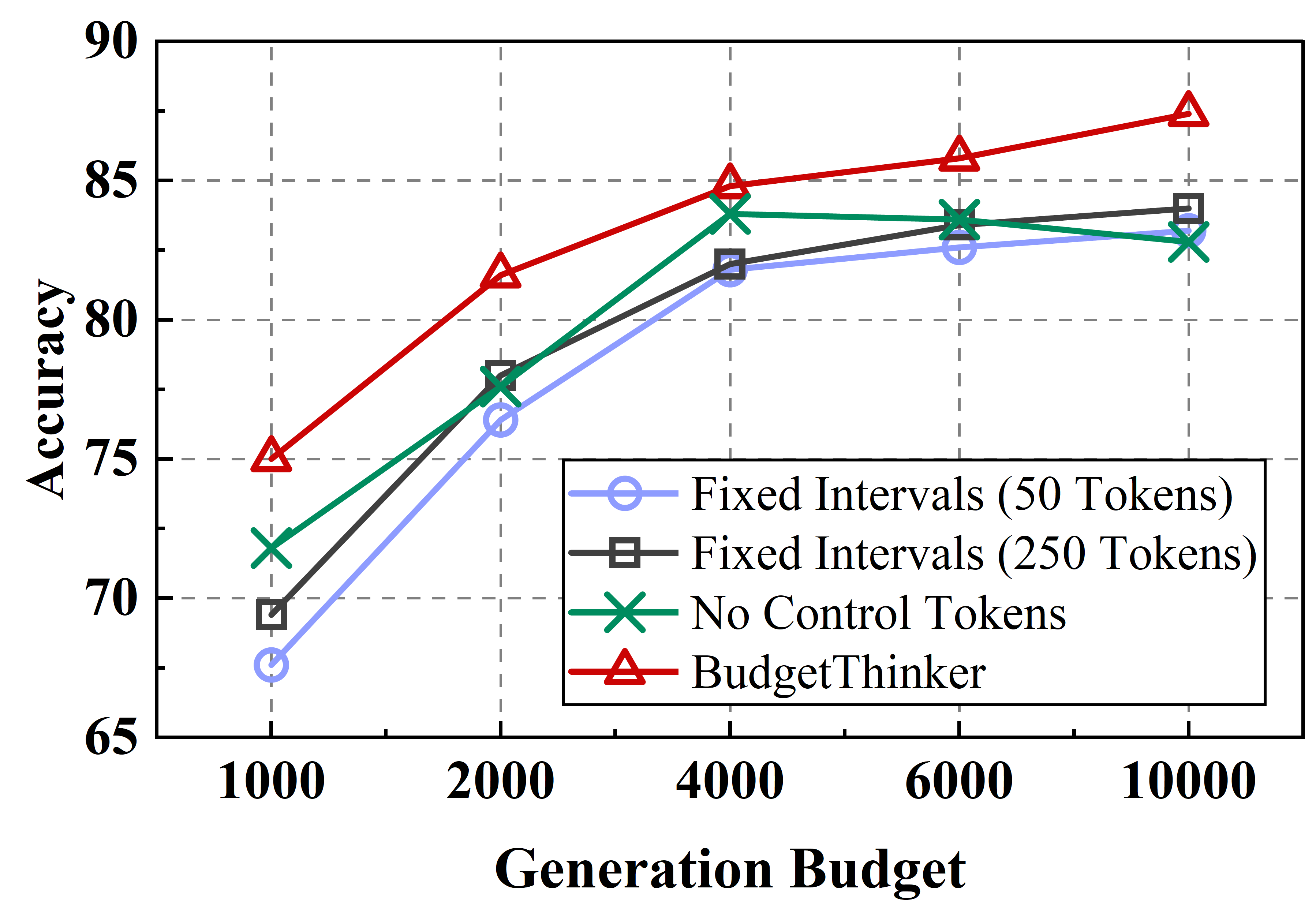}
        \caption{Analysis of control token insertion strategies.}
        \label{fig:control_tokens}
    \end{subfigure}
    \hspace{0.0025\textwidth}
    \begin{subfigure}{0.435\textwidth}
        \centering
        \includegraphics[width=\linewidth]{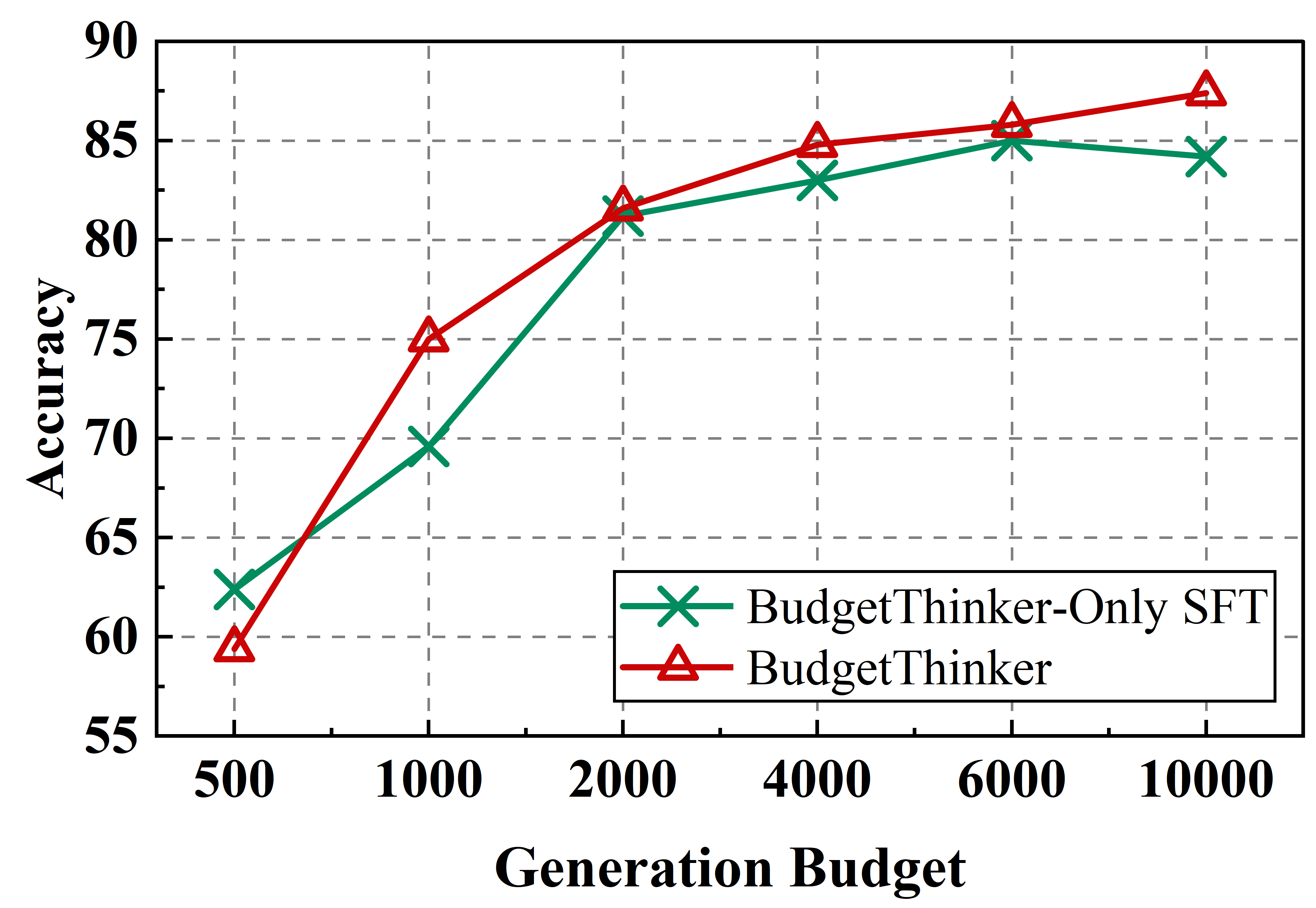}
        \caption{Reinforcement learning ablation. }
        \label{fig:rl_ablation}
    \end{subfigure}
    \caption{Ablation studies on control strategy and RL training (MATH-500, 1.5B Model).}
    \label{fig:main}
\end{figure}

We analyze the performance of \name under four distinct control token insertion strategies to understand their impact:
\begin{itemize}
    \item \textbf{No Control Tokens:} An ablation where no special tokens are inserted, relying solely on the RL objective for budget control.
    \item \textbf{Fixed Interval (50 tokens):} Control tokens are inserted at a fixed, frequent interval of every 50 tokens, with 200 special tokens added for a maximum budget of 10000 tokens. 
    \item \textbf{Fixed Interval (250 tokens):} Control tokens are inserted at a larger, fixed interval of every 250 tokens, with 40 special tokens added for a maximum budget of 10000 tokens. 
    \item \textbf{Budget Ratio (Default):} Our proposed method, where 8 tokens are inserted at intervals corresponding to each 1/8th of the total budget.
\end{itemize}

Note that while the ``No Control Tokens'' and ``Budget Ratio'' strategies are inherently scalable to any context length, fixed-interval strategies require adjustments to the number of special tokens for different budget sizes.

The results, presented in Figure \ref{fig:control_tokens}, indicate that the \textbf{Budget Ratio} strategy yields the best performance. This suggests that a relative, ratio-based notification of remaining resources is more intuitive for the LLM to learn and adapt its reasoning pace, compared to fixed-interval notifications. Furthermore, the 250-token interval outperforms the 50-token interval, implying that a sparser, less intrusive set of control signals is more beneficial. The superior performance of our default strategy over the ``No Control Tokens'' baseline further validates the effectiveness of our core design. 

% \subsection{Ablation Study}
% \label{sec:ablation}

\subsection{Analysis of Iterative Training}
\label{sec:exp_iterative}

\begin{table}[ht]
\centering
\small
% 为了美观，稍微增加行高
\begin{tabular}{l cccc cccc}
\toprule
\multirow{2}{*}{Model} & \multicolumn{4}{c}{Accuracy (\%)} & \multicolumn{4}{c}{Budget Following Ratio (\%)} \\
\cmidrule(lr){2-5} \cmidrule(lr){6-9}
& $B=2K$ & $B=4K$ & $B=6K$ & $B=10K$ & $B=2K$ & $B=4K$ & $B=6K$ & $B=10K$ \\
\midrule
\name 6k & \cellcolor{blue!5!white} 79.6 & \cellcolor{blue!36!white} 85.2 & \cellcolor{blue!60!white} 86.0 & \cellcolor{blue!5!white} 84.6 & \cellcolor{blue!5!white} 47.0 & \cellcolor{blue!5!white} 67.8 & \cellcolor{blue!5!white} 76.2 & \cellcolor{blue!6!white} 85.6 \\
\name 6k-4k & \cellcolor{blue!11!white} 80.0 & \cellcolor{blue!60!white} 86.0 & \cellcolor{blue!5!white} 84.6 & \cellcolor{blue!40!white} 86.4 & \cellcolor{blue!27!white} 62.8 & \cellcolor{blue!27!white} 78.6 & \cellcolor{blue!30!white} 85.8 & \cellcolor{blue!40!white} 93.8 \\
\name 6k-4k-3k & \cellcolor{blue!11!white} 80.0 & \cellcolor{blue!5!white} 84.2 & \cellcolor{blue!21!white} 85.0 & \cellcolor{blue!56!white} 87.2 & \cellcolor{blue!32!white} 66.4 & \cellcolor{blue!39!white} 84.6 & \cellcolor{blue!41!white} 89.8 & \cellcolor{blue!47!white} 95.4 \\
\name 6k-4k-3k-2k & \cellcolor{blue!60!white} 83.6 & \cellcolor{blue!17!white} 84.6 & \cellcolor{blue!5!white} 84.6 & \cellcolor{blue!17!white} 85.2 & \cellcolor{blue!60!white} 86.8 & \cellcolor{blue!60!white} 95.4 & \cellcolor{blue!60!white} 97.2 & \cellcolor{blue!60!white} 98.6 \\
\midrule
\name (Full Training) & \cellcolor{blue!33!white} 81.6 & \cellcolor{blue!23!white} 84.8 & \cellcolor{blue!52!white} 85.8 & \cellcolor{blue!60!white} 87.4 & \cellcolor{blue!50!white} 79.8 & \cellcolor{blue!60!white} 95.4 & \cellcolor{blue!54!white} 95.0 & \cellcolor{blue!54!white} 97.2 \\
\bottomrule
\end{tabular}
% \caption{Iterative RL training analysis. The table shows model accuracy (\%) and budget following ratio of \name-1.5B on MATH-500 dataset under different budgets. Darker blue corresponds to larger values within the same budget. ``6k'', ``6k-4k'' denote training only on 6k generation budget and training iteratively by 6k budget and 4k budget. Full training denotes randomly sampling budgets after training on 6k$\sim$2k budgets. }
\caption{This table tracks the Accuracy (\%) and Budget Following Ratio (\%) of \name-1.5B at different stages of the iterative training. Each row represents a checkpoint (\eg, ``6k-4k'' denotes the model after training sequentially on 6k and 4k budgets). The ``Full Training'' model completes the 6k-to-2k curriculum plus a final mixed-budget training phase where budgets are randomly sampled. Darker cells indicate higher values within each column.}
\label{tab:iterative_training_ablation}

\end{table}

To investigate the impact of our iterative training curriculum, we evaluated checkpoints of \name-1.5B on MATH-500, as shown in Table~\ref{tab:iterative_training_ablation}. The results reveal that as the model is trained on smaller budgets, its reasoning capability on larger budgets initially decreases. However, after the final mixed-budget training stage, accuracy on larger budgets recovers. We also observed that training on smaller budgets improves the budget following ratio, encouraging the model to generate more concise answers. After the full curriculum, the LLM achieves a balanced capability to handle all budgets, even if it is not individually optimal for every single budget constraint.
% To investigate the impact of our iterative training curriculum, we compare it against a model trained with budgets randomly sampled from $B \in \{2000, 4000, 6000\}$ from the outset. As shown in Figure~\xx, % TODO: Add figure for iterative training ablation
% the model trained without the curriculum exhibits limited performance gains, highlighting the crucial role of iterative training. Our curriculum, which starts with larger budgets and progressively reduces them, allows the model to first learn unconstrained reasoning before adapting to limitations. The final mixed-budget finetuning phase then harmonizes its capabilities across the entire spectrum of budgets, achieving a superior trade-off compared to the random sampling approach.

\subsection{Ablation on Reinforcement Learning}
\label{sec:exp_rl}
To isolate the contribution of reinforcement learning, we compare the full \name model with a version trained only with Supervised Fine-Tuning (SFT). The results in Figure \ref{fig:rl_ablation} % TODO: Add figure for RL ablation
demonstrate that the full RL-tuned model consistently surpasses the SFT-only model across all tested budgets. This underscores the importance of the RL phase. While SFT teaches the model the format and semantic meaning of control tokens, RL encourages the model to actively explore and discover more effective reasoning strategies that optimize the reward under specific budget constraints, ultimately leading to higher accuracy.

\section{Related Work}

\subsection{Test-Time Scaling for Reasoning LLMs}
Recent advances in test-time scaling seek to improve LLM reasoning by increasing computational depth during decoding. 
Among these test-time-scaling methods, reinforcement learning encourages LLMs to explore different strategies of solving problems and allocate more reasoning times for reflection. It has shown great potential in boosting LLM reasoning in multiple domains, including math \citep{deepseekr1, gemini25}, coding \citep{hui2024qwen2, yang2025qwen3technicalreport}, agentic tasks \citep{kimiteam2025kimik2openagentic}, and multimodal reasoning \citep{reasonrft, vlmr1}. 
Some other works propose to distill reasoning ability from long reasoning CoTs generated by large models to smaller LLMs to encourage deep thinking \cite{muennighoff2025s1, bespoke_stratos, sky_t1_2025, ye2025limo, xu2025redstardoesscalinglongcot, latent_reasoning}. 
While effective at enhancing complex problem-solving, these methods often suffer from significant inference latency due to the generation of lengthy outputs \citep{Sun2025TimesUp, zhu2025conciseadaptivethinkinglarge, meta_rft}, rendering them impractical for deployment in real-time systems \citep{adaptivenet, flexnn, legodnn}. 
Besides, recent works also found that lengthy outputs often result in overthinking (\eg meaningless repeats) and even lead to errors \citep{yu2025dapo, chen2025think23overthinkingo1like}. Our method teaches LLMs how to follow a user-specified token budget during its reasoning process. This helps the model generate more controlled and efficient reasoning.

\subsection{Efficient Chain of Thoughts}

% \wh{list the works here: https://github.com/Hongcheng-Gao/Awesome-Long2short-on-LRMs/}

Several methods have been proposed to alleviate the token efficiency problem of LLMs' reasoning. 
Prompt methods \citep{xu2025cod,tokenaware,muennighoff2025s1} make LLMs generate less reasoning tokens directly by adding explicit length constraints into the prompts. Solution routing works \citep{yu2025thinksmarterharderadaptive,wang2025stepwiseinformativenesssearchefficient} allow for mid-generation control to prune unpromising traces. Computation routing methods \citep{damani2024learninghardthinkinputadaptive, fu2025efficientlyscalingllmreasoning,wang2025makepennycountdifficultyadaptive, li2025selfbudgeter} choose to allocate just the necessary computation budget based on predicted complexity of queries, either by routing to larger models or conducting more samplings to vote for the final answer. 
There are also learning methods that either construct well-designed datasets to apply model fine-tuning \citep{chen2025think23overthinkingo1like,tokenaware,xia2025tokenskipcontrollablechainofthoughtcompression,zeng2025pruningunsurprisingefficientcode} or use RL methods \citep{yu2025dapo,yang2025qwen3technicalreport,l1,hou2025thinkprunepruninglongchainofthought,arora2025training} to encourage model to generate concise yet accurate answers.
More recently, models like Qwen3 \citep{yang2025qwen3technicalreport}, Kimi k1.5 \citep{kimiteam2025kimik15scalingreinforcement} and GPT-5 \citep{openai2025gpt5} integrate hybrid thinking modes of long and short CoTs, seeking trade-off between output length and model performance. Instead, our work achieves precise token budget control. 
LLMs are trained to understand and follow the constraint in prompts and adaptively adjust their reasoning process to complete the task within the user-requested budgets, showcasing the flexibility and efficiency of our method.
\section{Conclusion}

In this work, we introduce \name, a novel framework that manages the trade-off between reasoning quality and computational cost in LLMs. By using special control tokens and a two-stage training pipeline, \name enables precise, budget-aware control over the model's reasoning length. Our experiments validate that this method achieves superior budget adherence while maintaining high accuracy on challenging benchmarks. Thus, \name represents a significant step towards developing more efficient LLMs suitable for real-time, resource-constrained applications.

\bibliography{iclr2025_conference}
\bibliographystyle{iclr2025_conference}

% \input{tex/appendix}

% \appendix
% \section{Appendix}

\end{document}